\RequirePackage{snapshot}
\documentclass[final]{cvpr}

\usepackage{times}
\usepackage{epsfig}
\usepackage{graphicx}
\usepackage{amsmath}
\usepackage{amssymb}
\usepackage{makecell}

\usepackage{booktabs}
\usepackage{breqn}
\usepackage{mathtools}
\usepackage{braket}
\usepackage{placeins}
\usepackage{capt-of} %

\usepackage[pagebackref=true,breaklinks=true,colorlinks,bookmarks=false]{hyperref}

\newcommand{\nullt}{\textit{null}}
\newcommand{\separatort}{\textit{separator}}
\newcommand{\nopositions}{n_{positions}}
\newcommand{\notokens}{n_{tokens}}
\newcommand{\nomaxobjects}{n_{max}}
\newcommand{\vqgan}{VQGAN}

\newcommand{\RR}{\mathbb{R}}
\newcommand{\codebook}{\mathcal{Z}}
\newcommand{\discriminator}{D}
\newcommand{\decoder}{G}
\newcommand{\encoder}{E}

\makeatletter
\def\@eqnnum{{\normalsize \normalcolor (\theequation)}}
\makeatother

\def\cvprPaperID{5}
\def\confYear{CVPR 2021}

\begin{document}

\newcommand{\figuremethod}        {
   \begin{figure*}
   \begin{center}
   \includegraphics[width=0.95\linewidth]{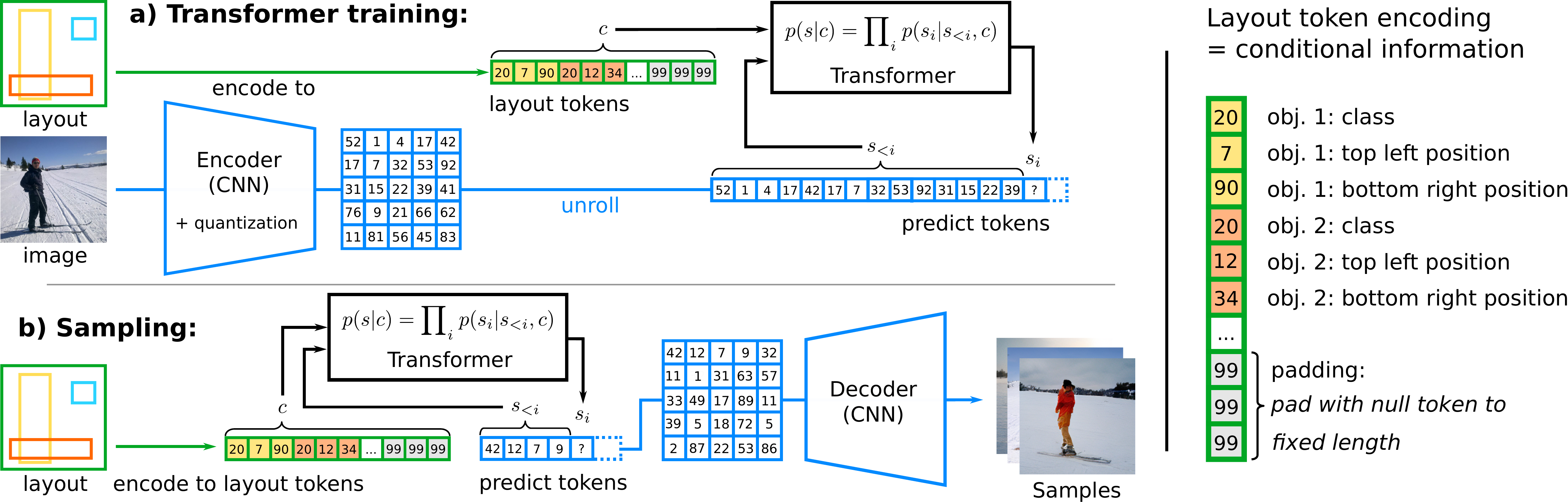}

   \end{center}
      \caption{
          Method.
          Encoder and decoder are trained in a first stage as a VQGAN~\cite{taming_vqgan} and encode image
          content in a compact, discrete latent space.
          \textbf{a)}~An autoregressive transformer is then trained to model the distribution of tokens within the latent space conditioned on object layouts.
          These layouts are received in the tokenized fashion shown on the right and are simply prepended to the image tokens.
          \textbf{b)}~After training, the codebook vectors sampled  by the transformer are then passed through the decoder for image rendering.
      }
   \label{fig:method}
   \end{figure*}
}

\newcommand{\firstpagefigure}{
  \vspace{-2.5em}
    \begin{center}
        \includegraphics[width=1.00\textwidth, trim=0em 0em 0em 0em, clip]{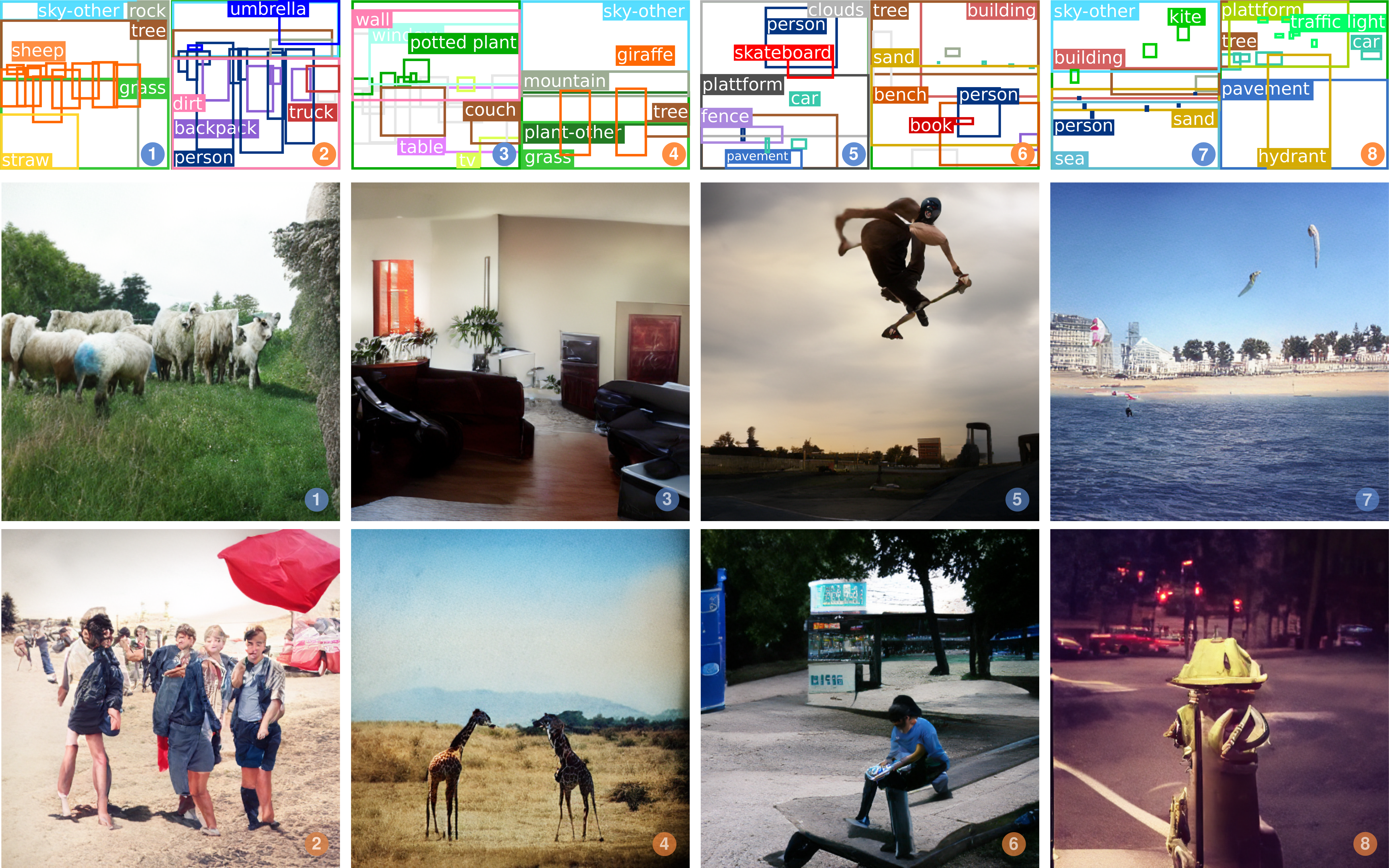}
      \captionof{figure}{
        From houseplants to fire hydrants, from skaters to kite surfers, from sheep to giraffes:
        Our model knows a broad range of objects and, given a bounding box layout, generates high-resolution images.
        It adds realistic context and details (for instance, short beachwear or marked sheep)
        and relates them in a meaningful way (for instance, skaters grabbing their board or people reading books close to them).
        These 512$\times$512\,pixel samples from our model trained on COCO data~\cite{caesar_2018_cocostuff} are best viewed zoomed in.}
      \label{fig:firstpagefigure}
    \end{center}
  \vspace{0.5em}
}

\newcommand{\figurecomparisonlostganlarge}        {
    \begin{figure}[t!]
    \begin{center}
        \includegraphics[width=1.0\linewidth]{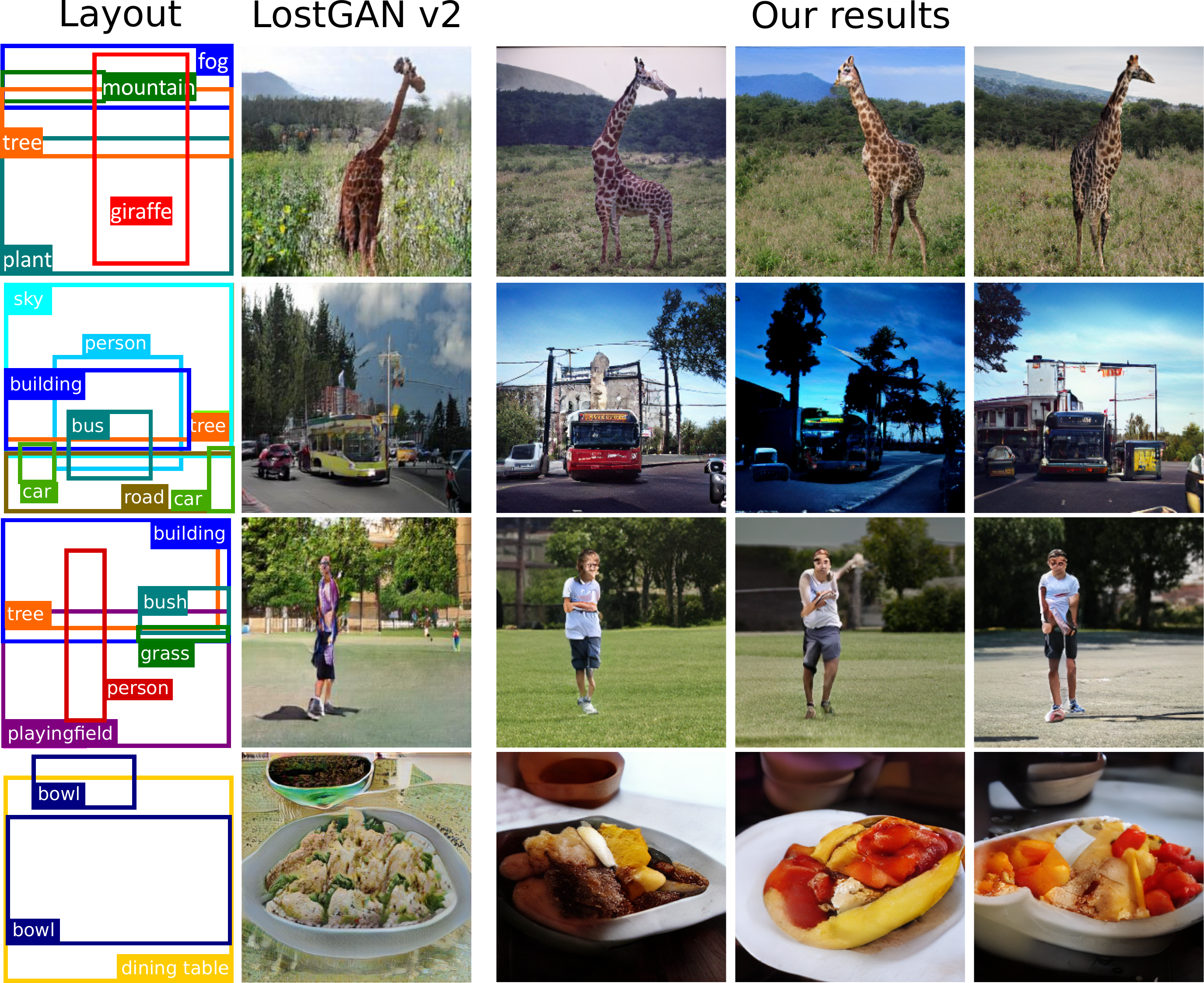}
    \end{center}
       \caption{
           Qualitative comparison on 256 and 512\,px with the SOTA model
           LostGAN-v2~\cite{sun_2020_learning_lost_gan_v_2}.
           Our model produces images with higher visual quality and consistency.
           Best viewed zoomed in.
       }
    \label{fig:comparison-lostgan-large}
    \end{figure}
}

\newcommand{\figurecomparisonallmethods}        {
    \begin{figure*}[t]
    \begin{center}
        \includegraphics[width=0.75\linewidth]{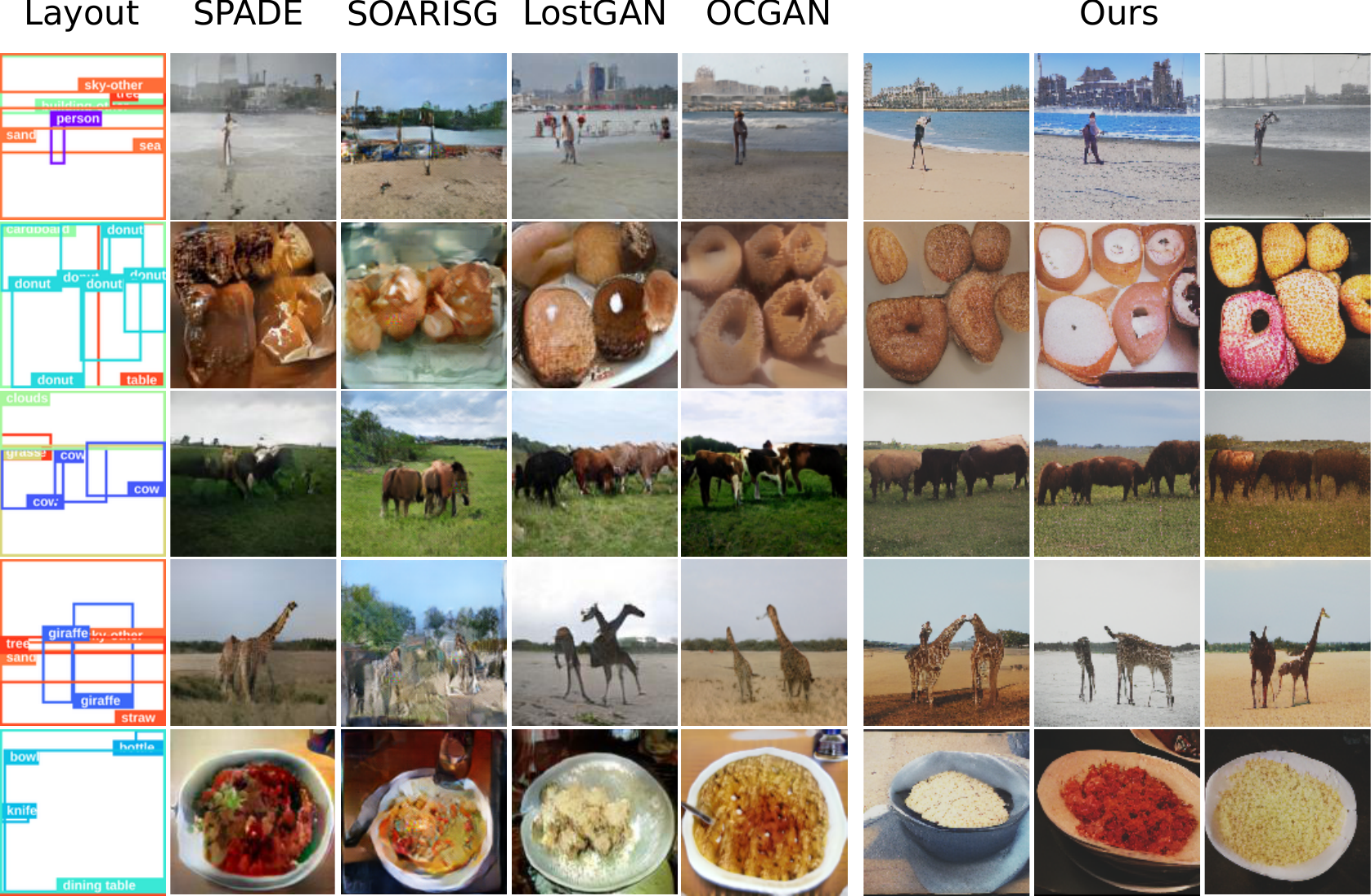}
    \end{center}
       \caption{
           Qualitative comparison with existing architectures for 128\,px.
           Our model produces consistent images with objects in the desired positions and high visual quality.
           We especially highlight the ability to produce sharp textures.
           Comparison model samples taken from~\cite{sylvain_oc_gan}.
       }
    \label{fig:comparison-all-methods-small}
    \end{figure*}
}

\newcommand{\figurezeroshotimages}        {
   \begin{figure*}
   \begin{center}
   \includegraphics[width=1.0\linewidth]{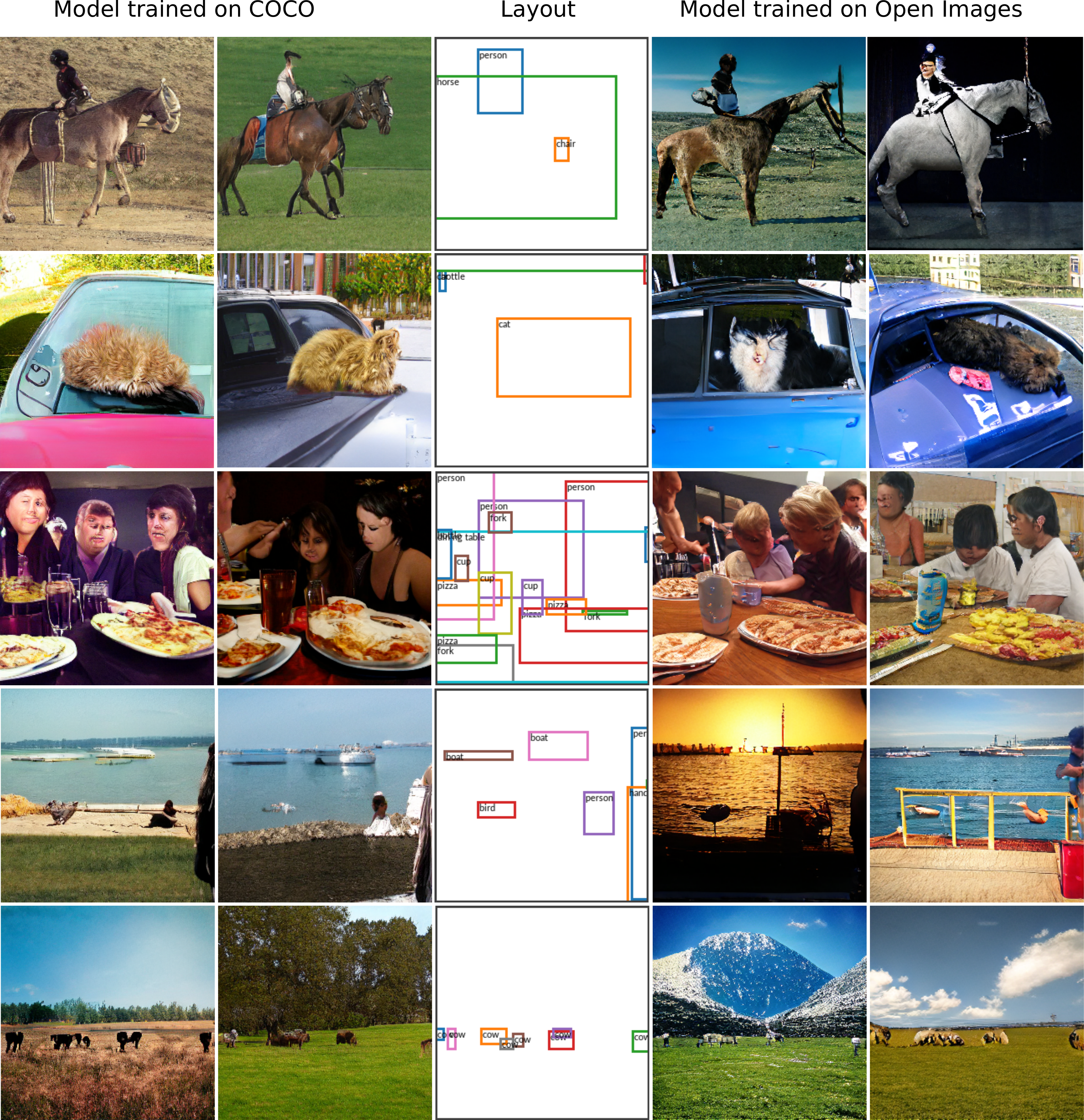}
   \end{center}
      \caption{
        Zero-shot scene image generation.
        We sample COCO test layouts with a model trained on full COCO (only \textit{things} classes)  and on Open Images on 256$\times$256\,px (330 classes).
        In order to do so, we filter the layouts for the 80 \textit{things} classes that are known to both our models.
        The COCO model is specialized on this dataset and thus presents a higher visual quality.
        The OpenImages model (not fully converged) generalizes well to COCO layouts and demonstrates a higher diversity.
      }
   \label{fig:zero-shot-images}
   \end{figure*}
}

\newcommand{\figurecomparisonlostgan}        {
   \begin{figure*}
   \begin{center}
   \includegraphics[width=0.97\linewidth]{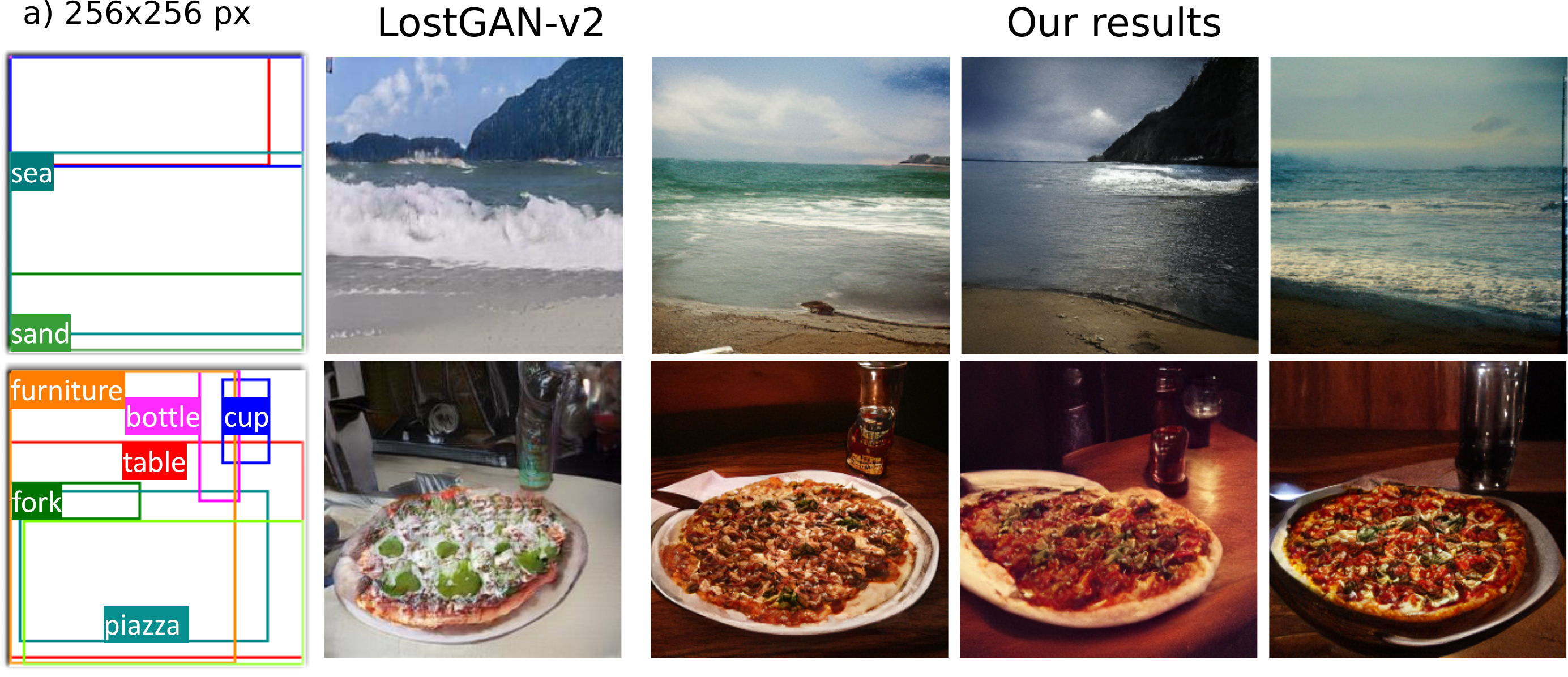}
   \includegraphics[width=0.97\linewidth]{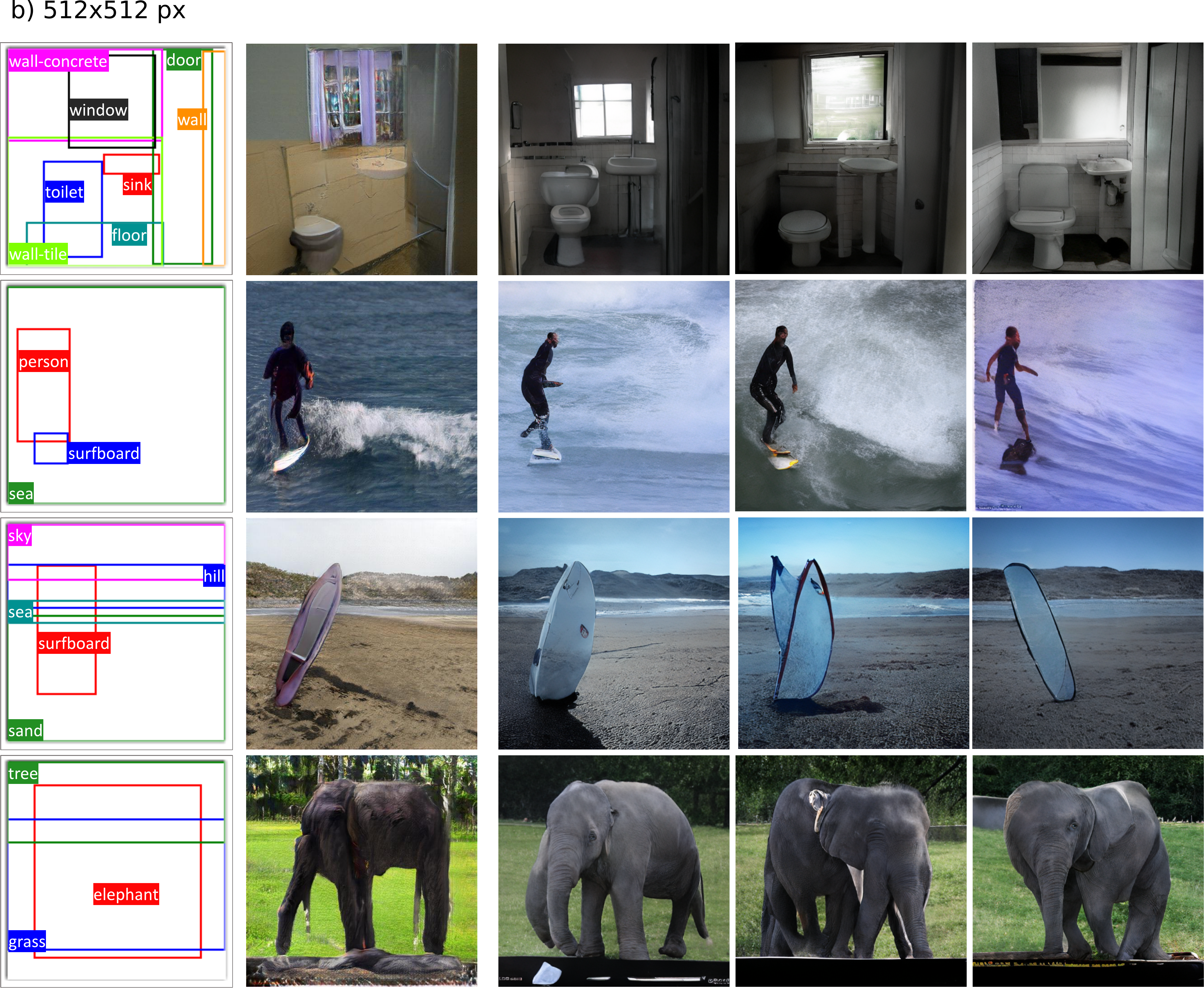}
   \end{center}
      \caption{
        Comparison with state-of-the-art method LostGAN-v2 COCO with resolutions of 256 and 512\,px.
        Comparison images taken from the original publication~\cite{sun_2020_learning_lost_gan_v_2}.
        Our model produces diverse images with high visual quality.
        It especially outperforms LostGAN-v2 on producing sharp textures, an ability inherited from
        its first-stage VQGAN.
        In general, image consistency is improved, even though producing the right amount of legs remains an open challenge.
      }
   \label{fig:comparison-lost-gan}
   \end{figure*}
}

\newcommand{\figurereconstruction}        {
   \begin{figure*}
   \begin{center}
   \includegraphics[width=0.75\linewidth]{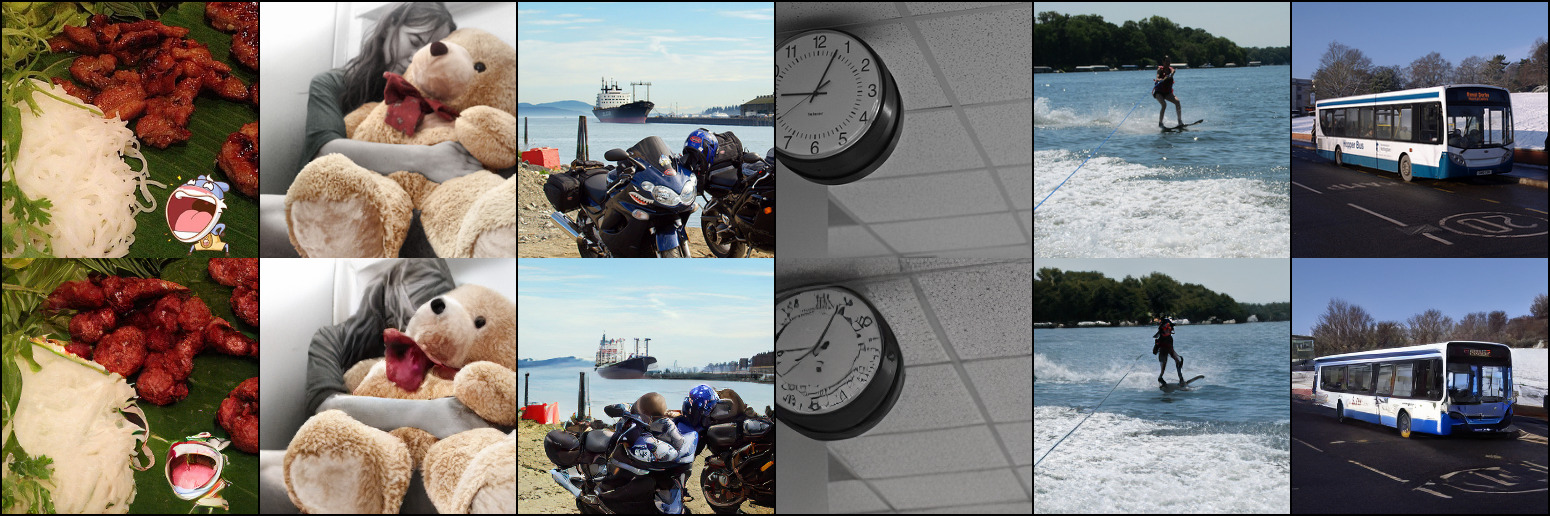}
   \end{center}
      \caption{
        These samples from Visual Genome illustrate the first-stage model's capacity for reconstructing images.
        Its strength are textures and an overall photorealistic look, more challenging is the reconstruction of faces and fine details, such as writing.
      }
   \label{fig:reconstruction}
   \end{figure*}
}

\newcommand{\figuresamplehalf}        {
   \begin{figure*}
   \begin{center}
   \includegraphics[width=1.0\linewidth]{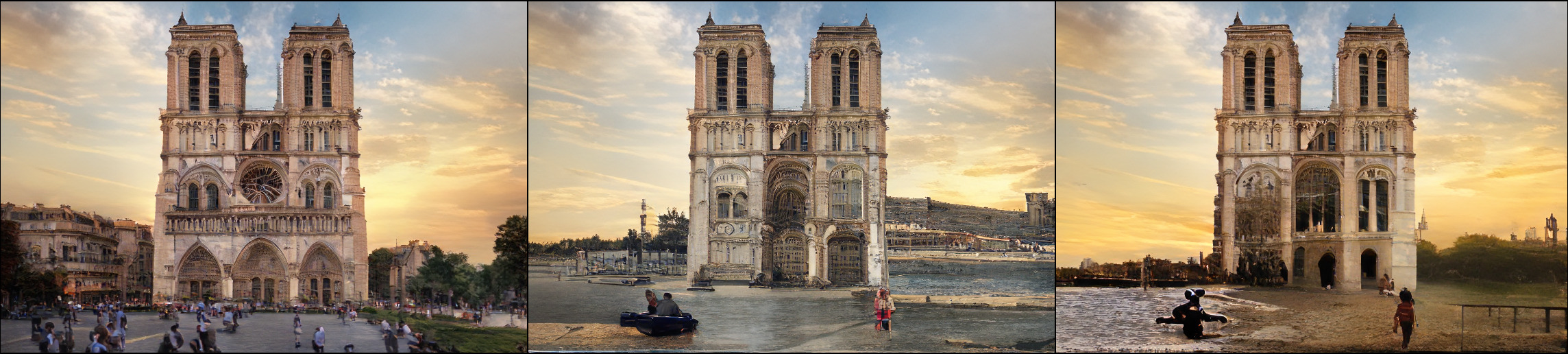}
   \end{center}
      \caption{
        Reimagining iconic scenes:
        The autoregressive model allows completing input images, \eg sampling the lower half when the upper half is given.
        Our model trained on Open Images re-imagines this famous view on the Notre Dame Cathedral (left: our first-stage's reconstruction of the original)
        and by adding surfers and boats, we place Notre Dame on the beach.
        Note how Notre Dame's Gothic architecture influences the overall image style.
        These images were sampled on a resolution of 512$\times$768\,px.
      }
   \label{fig:notre-dame-completion}
   \end{figure*}
}

\newcommand{\notavail} {N/A} %

\newcommand{\cococomparison}{

\begin{table}
\begin{footnotesize}
\begin{center}
\begin{tabular}{c c c c}
& Method & FID & SceneFID\\
\toprule

\makecell{128$\times$128$^{\dagger}$} &
\makecell{
   LostGAN$^\ast$~\cite{sun_2019_lost_gan} \\
   LostGAN-V2$^\ast$~\cite{sun_2020_learning_lost_gan_v_2} \\
   OC-GAN$^\ast$~\cite{sylvain_oc_gan} \\
   \textit{ours}$^\ast$
} &
\makecell{
   29.65 \\
   \textbf{24.76} \\
   36.31 \\
   \textit{54.91}  %
} &
\makecell{
   20.03 \\
   \notavail \\
   16.76 \\
   \textit{26.01}  %
}
\\
\midrule

\makecell{128$\times$128$^{\ddagger}$} &
\textit{ours$^\ast$ / ours $^{\ast\ast}$} &
\textit{40.26 / 34.67} &  %
\textit{17.89 / \textbf{13.98}}

\\ \midrule

\makecell{256$\times$256$^{\dagger}$} &
\makecell{
   LostGAN-V2$^\ast$~\cite{sun_2020_learning_lost_gan_v_2} \\
   OC-GAN$^\ast$~\cite{sylvain_oc_gan} \\
   SPADE$^{\diamond\ast}$~\cite{park2019semantic,sun_2020_learning_lost_gan_v_2} \\
   \textit{ours}$^\ast$
} &
\makecell{
   42.55 \\
   41.65 \\
   41.11 \\
   \textit{56.58}    %
} &
\makecell{
   \notavail \\
   \notavail \\
   \notavail \\
   \textit{24.07}
}
\\ \midrule

\makecell{256$\times$256$^{\ddagger}$} &
\textit{ours$^\ast$ / ours$^{\ast\ast}$} &
\textit{38.57 / \textbf{33.68}} & 
\textit{15.74 / \textbf{13.36}}

\\ \bottomrule

\end{tabular}
\end{center}
\caption{
   Comparison with current SOTA methods on $^\ast$Segmentation COCO (25k training images) and
   $^{\ast\ast}$full COCO (74k images, see Sec.~\ref{subsec:app-datasets}).
   Augmentation: $^{\dagger}$just flips, $^{\ddagger}$flips and random crops along longer axis.
   Our results back the finding that autoregressive models increase performance with higher data
   amount~\cite{henighan_2020_scaling,kaplan_2020_scaling} and indicate that this is also true for stronger
   data augmentation, \ie a virtual increase of data. \\
   $^\diamond$Using \cite{sun_2020_learning_lost_gan_v_2}'s coarse-layout-to-image setup for comparability.
}
\label{tab:cococomparison}
\end{footnotesize}
\end{table}
}

\newcommand{\vgcomparison}{
\begin{table}
\begin{footnotesize}
\begin{center}
\begin{tabular}{c c c c}
& Method & FID & SceneFID \\
\toprule

\makecell{128$\times$128$^{\dagger}$} &
\makecell{
   LostGAN~\cite{sun_2019_lost_gan} \\
   LostGAN-V2~\cite{sun_2020_learning_lost_gan_v_2} \\
   OC-GAN~\cite{sylvain_oc_gan} \\
   \textit{ours}
} &
\makecell{
   29.36 \\
   29.00 \\
   28.26 \\
   \textit{28.96}  %
} &
\makecell{
   \notavail \\
   \notavail \\
   \textbf{9.63} \\
   \textit{12.78}
}

\\ \midrule

\makecell{128$\times$128$^{\ddagger}$} &
\textit{ours} &
\textbf{\textit{21.33}} &  %
\textit{11.14}

\\ \midrule

\makecell{256$\times$256$^{\dagger}$} &
\makecell{
   LostGAN-V2~\cite{sun_2020_learning_lost_gan_v_2} \\
   OC-GAN~\cite{sylvain_oc_gan} \\
   \textit{ours}
} &
\makecell{
   47.62 \\
   40.85 \\
   \textit{28.05}   %
} &
\makecell{
   \notavail \\
   \notavail \\
   \textit{11.35}
}

\\ \midrule

\makecell{256$\times$256$^{\ddagger}$} &
\textit{ours} &
\textbf{\textit{19.14}} &   %
\textbf{\textit{8.61}}

\\ \bottomrule

\end{tabular}
\end{center}
\caption{
   Comparison with current SOTA methods on the Visual Genome dataset.
   Due to the size of the dataset results are on-par with previous methods
   even with basic data augmentation.
   Augmentation as Tab.~\ref{tab:cococomparison}.
   Competitive scores as reported by~\cite{sylvain_oc_gan}.
}
\label{tab:vgcomparison}
\end{footnotesize}
\end{table}
}

\newcommand{\allcomparison}{
\begin{table}
\begin{footnotesize}
\begin{center}
\begin{tabular}{c c c c c c c}
& \multicolumn{3}{c}{FID}  & \multicolumn{3}{c}{SceneFID} \\
& COCO & VG & OI           & COCO & VG & OI         \\
\toprule

256$\times$256 &
\textit{32.76} &  %
\textit{18.40} &  %
\textit{45.33} &  %
\textit{8.98} &   %
\textit{6.95} &   %
\textit{15.85}    %

\\ \midrule

512$\times$512 &
\textit{39.64} &
\textit{45.73} &  %
\textit{48.11} &  %
\textit{21.69} &  %
\textit{15.27} &  %
\textit{17.97}    %

\\ \bottomrule

\end{tabular}
\end{center}
\caption{
   Solving a more difficult problem on higher resolution:
   Including images with 2--30 objects of any size.
   The resulting increase in training data and a transformer with more parameters allow us to keep pace
   with the new complexity and give even better results than in Tables~\ref{tab:cococomparison} and~\ref{tab:vgcomparison}.
   We outperform FID scores reported in~\cite{sun_2020_learning_lost_gan_v_2} by 13--60\%, even though the comparability
   suffers due to different object number.
   COCO samples are presented in Figure~\ref{fig:firstpagefigure}.
}
\label{tab:allcomparison}
\end{footnotesize}
\end{table}
}

\title{High-Resolution Complex Scene Synthesis with Transformers}

\author{Manuel Jahn$^\ast$ \and Robin Rombach \vspace{0.5em} \\
IWR, HCI, Heidelberg University
\and Bj\"orn Ommer
}

\twocolumn[{%
\maketitle%
\firstpagefigure%
}]

\renewcommand{\thefootnote}{\fnsymbol{footnote}}

\begin{abstract}
\enlargethispage{1\baselineskip}
\vspace{-0.5em}
The use of coarse-grained layouts for controllable synthesis of complex scene images via deep generative models has recently gained popularity.
However, results of current approaches still fall short of their promise of high-resolution synthesis.
We hypothesize that this is mostly due to the highly engineered nature of these approaches which often rely on auxiliary losses and intermediate steps such as mask generators.
In this note, we present an orthogonal approach to this task, where the generative model is based on pure likelihood training without additional objectives.
To do so, we first optimize a powerful compression model with adversarial training which learns to reconstruct its
inputs via a discrete latent bottleneck and thereby effectively strips the latent representation of high-frequency
details such as texture.
Subsequently, we train an autoregressive transformer model to learn the distribution of the discrete image
representations conditioned on a tokenized version of the layouts.
Our experiments show that the resulting system is able to synthesize high-quality images consistent with the given layouts.
\enlargethispage{2\baselineskip} In particular, we improve the state-of-the-art FID score on COCO-Stuff and on Visual Genome by up to $19\%$ and $53\%$
and demonstrate the synthesis of images up to 512$\times$512\,px on COCO and Open Images.
\footnotetext[1]{Corresponding author: \href{mailto:vp248@stud.uni-heidelberg.de}{vp248@stud.uni-heidelberg.de}}
\end{abstract}
\figuremethod

\section{Introduction}
\label{sec:intro}
\enlargethispage{1\baselineskip}
\noindent
Recent years have brought unprecedented advances to conditional image generation with neural networks, \eg in class-conditional~\cite{brock2018large, razavi2019generating}, text-conditional~\cite{ramesh_2021_dalle, attngan} or pixel-wise conditional settings~\cite{park2019semantic, wang2018pix2pixHD, oord2016conditional}.
To capture the inherent ambiguity of this task, typically \emph{generative} models are used, which learn the distribution of images conditioned on the given information.
A particular class of conditional models deals with synthesis of complex scenes from coarse layouts, often specified as scene graphs~\cite{johnson_2018_image, ashual_2019_soarisg, wolf_scene_graph_2020} or bounding boxes~\cite{sylvain_oc_gan, sun_2020_learning_lost_gan_v_2}.
In addition to learning the shape and appearance of single objects, these models are also explicitly forced to model the \emph{interactions} between objects. %
However, achieving high-resolution synthesis results of complex scenes remains a challenging task.
We hypothesize that this is partly due to the fact that most concurrent approaches rely on many steps and
auxiliary losses such as scene-matching or object-specific losses~\cite{johnson_2018_image, ashual_2019_soarisg, sylvain_oc_gan, wolf_scene_graph_2020},
which turns out limiting the models' overall performance.

\noindent In an orthogonal approach, \cite{taming_vqgan} and~\cite{ramesh_2021_dalle} consider a more generic mechanism for conditional image synthesis via training an autoregressive (AR) transformer model in the latent space of a pre-trained discrete autoencoder model~\cite{oord2018neural, jang2017gumbel}, where~\cite{taming_vqgan} achieve efficient 
high-resolution synthesis by extending the reconstruction task with an adversarial objective.
By building on~\cite{taming_vqgan}, this note demonstrates how such a transformer-based system can be applied to high-resolution scene synthesis from coarse layouts via pure AR likelihood training and \emph{with no additional training objectives}.
In particular, we improve FID by 19--53\% on the previous state of the art for 256$\times$256\,px. %
Moreover, we provide first experiments on layout-guided synthesis on the Open Images dataset~\cite{kuznetsova_2020_open_images} and confirm previous evidence~\cite{henighan_2020_scaling} regarding the scalability of autoregressive transformers.

\subsection{Related Work}
\enlargethispage{1\baselineskip}
\paragraph{Self-Attention and Transformers}
The basic \emph{transformer} model~\cite{vaswani2017attention} is a sequence-to-sequence model, which
models interactions between its elements through the \emph{attention mechanism}~\cite{BahdanauCB14,parikh2016decomposable}. This mechanism is \emph{global}, \ie interactions between sequence elements are computed irrespective of their relative positioning. Hence, the architecture does not
incorporate locality biases such as the convolution operation in CNNs~\cite{lecun1995convolutional, UlyanovVL17}.
\newline
\vspace{0.1em}
\noindent \textbf{VQGAN} extends previous work on two-stage generative modeling via discrete representation learning \cite{oord2018neural, razavi2019generating}
by using a more agressive downsampling scheme; enabled by adding an adversarial realism prior and a perceptual metric \cite{zhang2018perceptual} to the autoencoder objective in \cite{oord2018neural}.
Adversarial training of this model ensures (i) high-fidelity reconstructions and (ii) strong compression, and thereby provides a suitable space~\cite{SalimansK0K17, dieleman2020typicality, razavi2019generating, taming_vqgan} to train an autoregressive transformer-based likelihood model in a second stage.
We show how to extend this approach for scene image generation.

\vspace{-1.6em}
\paragraph{Layouts to Image}
Previous methods commonly rely on intermediate pixel-wise semantic label maps generated
via a mask regression (object shape) and a box regression network (object position)
trained with an L1 objective~\cite{johnson_2018_image, ashual_2019_soarisg} or an adversarial loss~\cite{wolf_scene_graph_2020}.
The final images are then produced via a series of convolutional ``refinement'' layers~\cite{johnson_2018_image, ashual_2019_soarisg}
or a pre-trained SPADE~\cite{park_2019_semantic_spade} model.
In contrast, \cite{sun_2019_lost_gan, sun_2020_learning_lost_gan_v_2, sylvain_oc_gan}~generate scenes in a GAN setting.
The generator is conditioned on the scene layout and mask predictors via a layout-aware norm, inspired by StyleGAN~\cite{karras_2019_stylegan}.
\cite{sylvain_oc_gan}~further introduce a differentiable graph-scene matching loss to enhance the separation of objects.
Orthogonal to these approaches, we avoid intermediate steps and after compressing high-frequency image details
(\eg textures) into our first stage, we model the data distribution in an abstract latent space, rendering auxiliary steps unnecessary.

\section{Method}
\label{sec:method}
\noindent
We aim to leverage the flexibility and expressivity of the transformer architecture and its attention mechanism to
generate high-resolution images from coarse layouts, while retaining the coherency of composition and interplay of
depicted objects through AR likelihood-based learning only. However, 
likelihood-based models have been demonstrated~\cite{SalimansK0K17} to overspend capacity on
short-range interactions of pixels when modeling images \emph{directly} in pixel space.
To resolve this problem,
we extend upon~\cite{taming_vqgan} and first utilize adversarial training to learn a discrete codebook $\mathcal{Z}$ in a low-dimensional space.
As a result, any image can be described by a \emph{short} sequence of integers (see Sec.~\ref{sec:vqgan}).
Given this codebook, we further follow~\cite{taming_vqgan} and train an AR transformer model on
the discrete image representation, conditioned on a suitable representation of the image layout
(see Sec.~\ref{subsec:conditional-encoding}).
Thus, the transformer model is trained in an abstract space where compositional-irrelevant details such
as high-frequency patterns are suppressed.
We then rely on the expressivity of the attention mechanism to model interactions of these abstract tokens and learn the \emph{composition} of the specified objects.
This approach, depicted in Fig.~\ref{fig:method}, is  able to generate realistic and consistent complex scene images.

\newcommand{\codebookdim}{n_z}

\subsection{Efficient Training in Latent Space}
\label{sec:vqgan}
\noindent\cite{taming_vqgan} demonstrated that adversarially guided reconstruction through
neural vector quantization~\cite{oord2018neural} enables learning a low-dimensional latent space of abstract,
discrete image tokens, 
suitable for downstream autoregressive MLE.
Their full model consists of a learnable codebook $\codebook$, an encoder $\encoder$, a decoder $\decoder$ and a
discriminator $\discriminator$, 
where the codebook contains $|\mathcal{\codebook}|$ learnable entries $\in \RR^{\codebookdim}$
indexed by integers $\{0, 1, \dots, |\mathcal{\codebook}|-1\}$.
After training, a sufficiently general model and corresponding $\mathcal{Z}$ can be reused across tasks.

\vspace{-1.45em}
\paragraph{Conditional Autoregressive Learning}
The learned codebook $\codebook$ and encoder $\encoder$ allow to represent any input image
$x \in \RR^{3 \times W \times H}$ as a sequence of integers $s(x) = s \in \{0, 1, \dots, |\mathcal{\codebook}|\}^{h \times w}$,
which index the learned codebook.
This sequence is obtained by quantization of the encoded image $\encoder(x) \in \RR^{w \times h}$.
$W \times H$ and $w \times h$ describe the spatial extent of input and latent representation, respectively.
After unrolling the sequence $s$ in a ``raster-scan'' order and given information  $c$
(\eg of a corresponding coarse layout), the transformer is trained to learn the likelihood of $s$ as
$ p(s \vert c) = \prod_i p(s_i \vert s_{<i}, c)$.
The image generation process is expressed as an autoregressive next-element prediction. %
Hence, the full objective for training the transformer model reads

\vspace{-1.2em}
{\small \begin{eqnarray}
  \mathcal{L} = - \mathbb{E}_{x, c\sim p(x, c)} \Big[ \log p\Big(s(x)\vert c\Big) \Big] .
\label{eq:transformerloss}
\end{eqnarray}}

\vspace{-0.6em}
\noindent Note that $c$ can be any information (i) associated with the data, as long as it is (ii) encoded in a suitable tokenized representation
 in order to apply the training objective in Eq.~\eqref{eq:transformerloss}.
Examples include class labels~\cite{razavi2019generating}, text~\cite{ramesh_2021_dalle} and or another model's representation~\cite{taming_vqgan}.
In the following section, we describe our choice of conditioning $c$.

\subsection{Encoding Coarse Layouts}
\label{subsec:conditional-encoding}
\vspace{-0.6em}
\noindent We consider scene image datasets that provide coarse layout information in form of bounding boxes.
We encode each object $y$ by a triple $o = (c_y, {tl}_y, {br}_y)$, with $c_y$ being the object class index,
$tl_y$ the top-left and $br_y$ the bottom-right corner position.
In order to encode positions, we put a virtual grid on our image and number grid intersections continuously from $0$ to $\nopositions$
(for details, see Sec.~\ref{subsec:app-encoding}).
We choose $\nopositions=\notokens-1$ and can cover all image positions in the image with little imprecision;
one token is reserved to serve as a \nullt{} token.
Furthermore, we opt for a fixed conditional sequence length and pad
with \nullt{} tokens when there are fewer than $\nomaxobjects$ objects.

\section{Experiments}
\label{sec:experiments}
\figurecomparisonlostganlarge
\noindent
This section provides our empirical investigation of the model introduced in Sec.~\ref{sec:intro} and~\ref{sec:method}.
First, we evaluate our model's ability to compete with state-of-the-art methods in layout-to-image on
COCO~\cite{caesar_2018_cocostuff} and Visual Genome~\cite{krishna_2016_visual_genome} on resolutions up to 256 pixels.
Next, we broaden the training exercise to enable higher scene complexity and include the Open Images
dataset~\cite{kuznetsova_2020_open_images} to sample images up to 512 pixels.
Finally, we demonstrate the model's ability to generalize in a ``zero-shot'' setting and sample COCO images with a
model exclusively trained on Open Images.

\vspace{-0.6em}
\paragraph{Architectures}
The architectural design of our models follows~\cite{taming_vqgan}:
For the \textbf{VQGAN}, we use the encoder-decoder structure as introduced in~\cite{ho2020denoising}
(without skip connections) and a patch-discriminator as in~\cite{wang2018pix2pixHD}.
The design of the \textbf{transformer} model follows the GPT-2 architecture~\cite{radford2019language_gpt2},
\ie blocks of multi-head self-attention, layer norm and point-wise MLPs. For choice of parameters, see Sec.~\ref{subsec:app-transformer}.

\cococomparison
\vgcomparison
\allcomparison

\vspace{-0.5em}
\subsection{Results on Complex Scene Synthesis}
\label{subsec:quant-results}
\enlargethispage{1\baselineskip}
\vspace{-0.2em}
\paragraph{COCO and Visual Genome}
Following standard practice, we use FID scores~\cite{heusel_2018_gans_fid} for a quantitative assessment of our approach.
We also note SceneFID when available, as~\cite{sylvain_oc_gan} suggests that FID is optimized for single-object images.
Results are reported in Table~\ref{tab:cococomparison},~\ref{tab:vgcomparison} and~\ref{tab:allcomparison}.

\noindent We observe that when strictly limiting data augmentation to horizontal flips as in previous works,
comparable FID and SceneFID scores can be reached for Visual Genome.
However, for the smaller Segmentation COCO dataset, one needs to add basic data augmentation like random cropping to achieve competitive performance.
A plus of this preprocessing is the model's ability to produce undistorted objects as we skip the resizing to a quadratic image.
Using full data augmentation for both COCO and Visual Genome (and using a data superset in the case of COCO),
we are able to outperform previous methods in the 256\,px setting by a margin of up to 18 FID points.

\vspace{-1.25em}
\paragraph{Increasing complexity}
We broaden the training exercise to include images with between 2 and 30 objects of any size.
More data becomes available as image filtering is less strict.
This benefits the autoregressive model and in conjunction with a larger transformer, even better FID and SceneFID scores
are reached in the 256\,px setting (Tab.~\ref{tab:allcomparison}).
Further, we use the sliding-window approach~\cite{taming_vqgan} to show that our model gives state-of-the-art FID scores
in a 512\,px setting.
Lastly, we report pioneer scores on the Open Images dataset.

\vspace{-1.25em}
\paragraph{Transfer to COCO}
Being roughly eleven times larger than COCO, Open Images is optimally suited for the autoregressive approach.
We demonstrate its power in a final step by rendering layouts of the COCO validation set with a model exclusively trained on
Open Images.
The visual quality of these renderings is high and the diversity immense, as one can see in the appendix (Figure~\ref{fig:zero-shot-images}).
Quantitatively, we get (Scene)FID scores of 36.4 (12.1) for the COCO-trained and 45.1 (14.0) for the OpenImages-trained model.

\section{Conclusion}
\label{sec:conclusion}
\enlargethispage{1\baselineskip}
\vspace{-0.3em}
\noindent We demonstrate that a conceptually simple model can outperform highly specialized systems when trained
in a suitable latent space: Guiding a VQ model through adversarial training (i)~yields
high-fidelity reconstructions while (ii)~providing an adequate space for downstream maximum-likelihood learning,
as the transformer modelling this space focuses on large-scale structure instead of short-range pixel interactions.
Conditioned on tokenized coarse layouts and without any additional
auxiliary losses the method outperforms previous state-of-the-art methods for complex scene generation and can be seamlessly
applied to high-resolution image generation.
After training on a large dataset (\textit{Open Images}),
our model can even achieve comparable performance on another dataset (\textit{COCO}) in a zero-shot setting.

{
    \small
    \bibliographystyle{ieee_fullname}
    \bibliography{ms}

\begin{thebibliography}{10}\itemsep=-1pt

\bibitem{krishna_2016_visual_genome}
{\em {Visual Genome: Connecting Language and Vision Using Crowdsourced Dense
  Image Annotations}}, 2016.

\bibitem{ashual_2019_soarisg}
Oron Ashual and Lior Wolf.
\newblock {Specifying Object Attributes and Relations in Interactive Scene
  Generation}.
\newblock In {\em 2019 {IEEE/CVF} International Conference on Computer Vision,
  {ICCV} 2019}, pages 4560--4568. {IEEE}, 2019.

\bibitem{BahdanauCB14}
Dzmitry Bahdanau, Kyunghyun Cho, and Yoshua Bengio.
\newblock {Neural Machine Translation by Jointly Learning to Align and
  Translate}.
\newblock In Yoshua Bengio and Yann LeCun, editors, {\em 3rd International
  Conference on Learning Representations, {ICLR} 2015}, 2015.

\bibitem{brock2018large}
Andrew Brock, Jeff Donahue, and Karen Simonyan.
\newblock {Large Scale GAN Training for High Fidelity Natural Image Synthesis}.
\newblock In {\em 7th International Conference on Learning Representations,
  {ICLR}}, 2019.

\bibitem{caesar_2018_cocostuff}
Holger Caesar, Jasper Uijlings, and Vittorio Ferrari.
\newblock {COCO-Stuff: Thing and Stuff Classes in Context}.
\newblock In {\em 2018 {IEEE} Conference on Computer Vision and Pattern
  Recognition, {CVPR} 2018}, pages 1209--1218. {IEEE} Computer Society, 2018.

\bibitem{dieleman2020typicality}
Sander Dieleman.
\newblock {Musings on typicality}, 2020.

\bibitem{taming_vqgan}
Patrick Esser, Robin Rombach, and Björn Ommer.
\newblock Taming transformers for high-resolution image synthesis.
\newblock {\em arXiv:2012.09841 [cs]}, 02 2021.

\bibitem{henighan_2020_scaling}
Tom Henighan, Jared Kaplan, Mor Katz, Mark Chen, Christopher Hesse, Jacob
  Jackson, Heewoo Jun, Tom~B. Brown, Prafulla Dhariwal, Scott Gray, Chris
  Hallacy, Benjamin Mann, Alec Radford, Aditya Ramesh, Nick Ryder, Daniel~M.
  Ziegler, John Schulman, Dario Amodei, and Sam McCandlish.
\newblock {Scaling Laws for Autoregressive Generative Modeling}.
\newblock {\em arXiv:2010.14701 [cs]}, 11 2020.

\bibitem{heusel_2018_gans_fid}
Martin Heusel, Hubert Ramsauer, Thomas Unterthiner, Bernhard Nessler, and Sepp
  Hochreiter.
\newblock {GANs Trained by a Two Time-Scale Update Rule Converge to a Local
  Nash Equilibrium}.
\newblock In Isabelle Guyon, Ulrike von Luxburg, Samy Bengio, Hanna~M. Wallach,
  Rob Fergus, S.~V.~N. Vishwanathan, and Roman Garnett, editors, {\em Advances
  in Neural Information Processing Systems 30: Annual Conference on Neural
  Information Processing Systems 2017}, pages 6626--6637, 2017.

\bibitem{ho2020denoising}
Jonathan Ho, Ajay Jain, and Pieter Abbeel.
\newblock {Denoising Diffusion Probabilistic Models}.
\newblock In Hugo Larochelle, Marc'Aurelio Ranzato, Raia Hadsell,
  Maria{-}Florina Balcan, and Hsuan{-}Tien Lin, editors, {\em Advances in
  Neural Information Processing Systems 33: Annual Conference on Neural
  Information Processing Systems 2020, NeurIPS 2020}, 2020.

\bibitem{wolf_scene_graph_2020}
Maor Ivgi, Yaniv Benny, Avichai Ben{-}David, Jonathan Berant, and Lior Wolf.
\newblock {Scene Graph to Image Generation with Contextualized Object Layout
  Refinement}.
\newblock {\em CoRR}, abs/2009.10939, 2020.

\bibitem{jang2017gumbel}
Eric Jang, Shixiang Gu, and Ben Poole.
\newblock {Categorical Reparameterization with Gumbel-Softmax}.
\newblock In {\em 5th International Conference on Learning Representations,
  {ICLR} 2017}. OpenReview.net, 2017.

\bibitem{johnson_2018_image}
Justin Johnson, Agrim Gupta, and Li Fei-Fei.
\newblock {Image Generation from Scene Graphs}.
\newblock In {\em 2018 {IEEE} Conference on Computer Vision and Pattern
  Recognition, {CVPR} 2018}, pages 1219--1228. {IEEE} Computer Society, 2018.

\bibitem{kaplan_2020_scaling}
Jared Kaplan, Sam McCandlish, Tom Henighan, Tom~B. Brown, Benjamin Chess, Rewon
  Child, Scott Gray, Alec Radford, Jeffrey Wu, and Dario Amodei.
\newblock {Scaling Laws for Neural Language Models}.
\newblock {\em arXiv:2001.08361 [cs, stat]}, 01 2020.

\bibitem{karras_2019_stylegan}
Tero Karras, Samuli Laine, and Timo Aila.
\newblock {A Style-Based Generator Architecture for Generative Adversarial
  Networks}.
\newblock In {\em {IEEE} Conference on Computer Vision and Pattern Recognition,
  {CVPR} 2019}, pages 4401--4410. Computer Vision Foundation / {IEEE}, 2019.

\bibitem{kuznetsova_2020_open_images}
Alina Kuznetsova, Hassan Rom, Neil Alldrin, Jasper Uijlings, Ivan Krasin, Jordi
  Pont-Tuset, Shahab Kamali, Stefan Popov, Matteo Malloci, Alexander
  Kolesnikov, Tom Duerig, and Vittorio Ferrari.
\newblock {The Open Images Dataset V4: Unified Image Classification, Object
  Detection, and Visual Relationship Detection at Scale}.
\newblock 2020.

\bibitem{lecun1995convolutional}
Yann LeCun, Yoshua Bengio, et~al.
\newblock {Convolutional Networks for Images, Speech, and Time Series}.
\newblock {\em The Handbook of Brain Theory and Neural Networks},
  3361(10):1995, 1995.

\bibitem{obukhov2020_torchfidelity}
Anton Obukhov, Maximilian Seitzer, Po-Wei Wu, Semen Zhydenko, Jonathan Kyl, and
  Elvis Yu-Jing Lin.
\newblock {toshas/torch-fidelity: Version 0.2.0}, May 2020.

\bibitem{parikh2016decomposable}
Ankur~P. Parikh, Oscar Täckström, Dipanjan Das, and Jakob Uszkoreit.
\newblock {A Decomposable Attention Model for Natural Language Inference}.
\newblock In Jian Su, Xavier Carreras, and Kevin Duh, editors, {\em Proceedings
  of the 2016 Conference on Empirical Methods in Natural Language Processing,
  {EMNLP} 2016}, pages 2249--2255. The Association for Computational
  Linguistics, 2016.

\bibitem{park2019semantic}
Taesung Park, Ming-Yu Liu, Ting-Chun Wang, and Jun-Yan Zhu.
\newblock {Semantic Image Synthesis with Spatially-Adaptive Normalization}.
\newblock In {\em Proceedings of the IEEE Conference on Computer Vision and
  Pattern Recognition, CVPR}, 2019.

\bibitem{park_2019_semantic_spade}
Taesung Park, Ming-Yu Liu, Ting-Chun Wang, and Jun-Yan Zhu.
\newblock {Semantic Image Synthesis with Spatially-Adaptive Normalization}.
\newblock In {\em {IEEE} Conference on Computer Vision and Pattern Recognition,
  {CVPR} 2019}, pages 2337--2346. Computer Vision Foundation / {IEEE}, 2019.

\bibitem{radford2019language_gpt2}
Alec Radford, Jeff Wu, Rewon Child, David Luan, Dario Amodei, and Ilya
  Sutskever.
\newblock {Language Models are Unsupervised Multitask Learners}.
\newblock 2019.

\bibitem{ramesh_2021_dalle}
Aditya Ramesh, Mikhail Pavlov, Gabriel Goh, Scott Gray, Chelsea Voss, Alec
  Radford, Mark Chen, and Ilya Sutskever.
\newblock {Zero-Shot Text-to-Image Generation}.
\newblock {\em arXiv:2102.12092 [cs]}, 02 2021.

\bibitem{razavi2019generating}
Ali Razavi, A{\"{a}}ron van~den Oord, and Oriol Vinyals.
\newblock {Generating Diverse High-Fidelity Images with VQ-VAE-2}.
\newblock In Hanna~M. Wallach, Hugo Larochelle, Alina Beygelzimer, Florence
  d'Alch{\'{e}}{-}Buc, Emily~B. Fox, and Roman Garnett, editors, {\em Advances
  in Neural Information Processing Systems 32: Annual Conference on Neural
  Information Processing Systems 2019, NeurIPS 2019}, pages 14837--14847, 2019.

\bibitem{SalimansK0K17}
Tim Salimans, Andrej Karpathy, Xi Chen, and Diederik~P. Kingma.
\newblock {PixelCNN++: Improving the PixelCNN with Discretized Logistic Mixture
  Likelihood and Other Modifications}.
\newblock In {\em 5th International Conference on Learning Representations,
  {ICLR} 2017, Toulon, France, April 24-26, 2017, Conference Track
  Proceedings}. OpenReview.net, 2017.

\bibitem{sun_2019_lost_gan}
Wei Sun and Tianfu Wu.
\newblock {Image Synthesis From Reconfigurable Layout and Style}.
\newblock In {\em 2019 {IEEE/CVF} International Conference on Computer Vision,
  {ICCV} 2019}, pages 10530--10539. {IEEE}, 2019.

\bibitem{sun_2020_learning_lost_gan_v_2}
Wei Sun and Tianfu Wu.
\newblock {Learning Layout and Style Reconfigurable GANs for Controllable Image
  Synthesis}.
\newblock {\em arXiv:2003.11571 [cs]}, 03 2020.

\bibitem{sylvain_oc_gan}
Tristan Sylvain, Pengchuan Zhang, Yoshua Bengio, R.~Devon Hjelm, and Shikhar
  Sharma.
\newblock {Object-Centric Image Generation from Layouts}.
\newblock {\em arXiv:2003.07449 [cs, eess]}, 12 2020.

\bibitem{UlyanovVL17}
Dmitry Ulyanov, Andrea Vedaldi, and Victor Lempitsky.
\newblock {Deep Image Prior}.
\newblock {\em arXiv:1711.10925}, 2017.

\bibitem{oord2016conditional}
Aaron van~den Oord, Nal Kalchbrenner, Oriol Vinyals, Lasse Espeholt, Alex
  Graves, and Koray Kavukcuoglu.
\newblock {Conditional Image Generation with PixelCNN Decoders}, 2016.

\bibitem{oord2018neural}
A{\"{a}}ron van~den Oord, Oriol Vinyals, and Koray Kavukcuoglu.
\newblock {Neural Discrete Representation Learning}.
\newblock In Isabelle Guyon, Ulrike von Luxburg, Samy Bengio, Hanna~M. Wallach,
  Rob Fergus, S.~V.~N. Vishwanathan, and Roman Garnett, editors, {\em Advances
  in Neural Information Processing Systems 30: Annual Conference on Neural
  Information Processing Systems 2017}, pages 6306--6315, 2017.

\bibitem{vaswani2017attention}
Ashish Vaswani, Noam Shazeer, Niki Parmar, Jakob Uszkoreit, Llion Jones,
  Aidan~N. Gomez, Lukasz Kaiser, and Illia Polosukhin.
\newblock {Attention is All you Need}.
\newblock In {\em Advances in Neural Information Processing Systems 30: Annual
  Conference on Neural Information Processing Systems, NeurIPS}, 2017.

\bibitem{wang2018pix2pixHD}
Ting-Chun Wang, Ming-Yu Liu, Jun-Yan Zhu, Andrew Tao, Jan Kautz, and Bryan
  Catanzaro.
\newblock {High-Resolution Image Synthesis and Semantic Manipulation with
  Conditional GANs}.
\newblock In {\em Proceedings of the IEEE Conference on Computer Vision and
  Pattern Recognition}, 2018.

\bibitem{attngan}
Tao Xu, Pengchuan Zhang, Qiuyuan Huang, Han Zhang, Zhe Gan, Xiaolei Huang, and
  Xiaodong He.
\newblock {AttnGAN: Fine-Grained Text to Image Generation With Attentional
  Generative Adversarial Networks}.
\newblock In {\em 2018 {IEEE} Conference on Computer Vision and Pattern
  Recognition, {CVPR} 2018}, pages 1316--1324. {IEEE} Computer Society, 2018.

\bibitem{zhang2018perceptual}
Richard Zhang, Phillip Isola, Alexei~A Efros, Eli Shechtman, and Oliver Wang.
\newblock The unreasonable effectiveness of deep features as a perceptual
  metric.
\newblock In {\em CVPR}, 2018.

\end{thebibliography}
}
\clearpage
\newpage

\appendix
\section{Appendix}
\label{sec:appendix}

\subsection{First stage: Details on VQGAN}
\label{subsec:app-vqgan}
\noindent Best results on all datasets for sizes $\geq 256$\,px are achieved with a \vqgan{} trained for 117 epochs on
256$\times$256\,px crops (along longer axis) of the full COCO dataset.
Our architecture uses four downsamplings yielding an internal representation of 16$\times$16 tokens.
The codebook size is 8192 tokens, each corresponding to a vector of dimensionality 256.

On 5k examples from the full COCO validation split, this model achieves a reconstruction FID (R-FID) of 11.0.
(compare to R-FID = 52.3 when using the DALL-E \cite{ramesh_2021_dalle} VQVAE instead). 

\noindent For the 128$\times$128\,px images we used an architecture with three downsamplings resulting in an internal
representation of 16$\times$16 and 16384 tokens with a dimensionality of 256.
This model is trained for 68 epochs on COCO data.

\subsection{Second stage: Details on Transformer}
\label{subsec:app-transformer}
\noindent We use the GPT-2 architecture~\cite{radford2019language_gpt2} in different sizes.
For the comparison with previous methods in Tables~\ref{tab:cococomparison} and~\ref{tab:vgcomparison} we use an
architecture with approximately 320 million parameters (24 layers, 16 attention heads, 1024 embedding dimensions and a dropout rate of 10\%).
For the broader and more difficult tasks addressed in Table~\ref{tab:allcomparison}, we use a more powerful architecture
with 1.1 billion parameters (36 layers and 1536 embedding dimensions, the remaining parameters identical).

\subsection{Details on Datasets}
\label{subsec:app-datasets}

\paragraph{Segmentation COCO~\cite{caesar_2018_cocostuff}}
For comparability, we evaluate the on the 2017 COCO Thing and Stuff
Segmentation Challenge\footnote{see \url{https://cocodataset.org/\#stuff-2017}} subset (50k images).
It features bounding boxes for 80 object and 91 \textit{stuff} classes (excluding ``other'').
We filter for images with 3 to 8 objects, each larger than 2~\% of the image's area.
We use the same split as~\cite{johnson_2018_image,sylvain_oc_gan} and obtain 24972 training, 1024 validation and 2048 test images.

\vspace{-1.0em}
\paragraph{Full COCO}
As autoregressive models greatly benefit of larger datasets~\cite{henighan_2020_scaling,kaplan_2020_scaling},
we make use of the full COCO Thing and Stuff dataset (164K images).
When filtering for 3 to 8 objects, one gets 74121 training images.
For comparability, validation is however done on the same 2048 test images as above.
Broadening the training exercise and filtering for 2 to 30 objects of any size, one obtains 112k training images.

\vspace{-1.0em}
\paragraph{Visual Genome~\cite{krishna_2016_visual_genome}}
We use the same filtering settings as~\cite{johnson_2018_image,sylvain_oc_gan} and obtain 62565 training,
5506 validation and 5088 test images with 178 object classes.
Broadening the training exercise, there are 80k training images.

\vspace{-1.0em}
\paragraph{Open Images V6~\cite{kuznetsova_2020_open_images}}
We promote the usage of this dataset in the field of scene image generation.
It contains 1.9 million images annotated with 16 million bounding boxes for 600 object classes.
This data richness (roughly eleven times larger than COCO) is highly beneficial for the data-demanding autoregressive approach.
We filter for the approximately 300 classes that have more than 1000 bounding boxes and further add
a handful of classes to increase compatibility with the COCO dataset (331 classes).
Filtering for images containing between 2 and 30 objects, we obtain 1.29 million training and 25k test images.

Visual Genome and Open Images differ from the COCO dataset in their focus on objects and in their absence of
texture and material classes (``stuff'': \textit{straw, sand, sky, \etc}).

\subsection{Details on Layout Encoding}
\label{subsec:app-encoding}
\noindent The first obvious layout encoding that comes to mind is done in five tokens: $o=(c_y, l_y, t_y, w_y, h_y)$.
However, since as the attention mechanism scales quadratically with sequence length, we employ the encoding
in three tokens by imposing a virtual grid on the image, with grid intersections numbered continuously in
``raster-scan'' order (\eg, the first row is encoded by: $0, 1, \dots, n_{col}=\lfloor \sqrt{\nopositions} \rfloor$).
The x-position can be computed from the position number $p$ as $x = p \bmod n_{col}$ and the y-position as $y = p//n_{col}$.

\noindent We further experimented with the introduction of a \separatort{} token in order to separate the object triples,
however, it was not found to be beneficial for model performance.
We omitted it to reduce sequence length.

\noindent In order to produce 512$\times$512\,px images, the sliding-window approach detailed in ~\cite{taming_vqgan} is used.
That means that the conditional information presented to the model needs to be adjusted in almost every step.
Our findings are that simple filtering for existing objects and rescaling bounding boxes to the new viewport gives
problematic results on large objects, since the transformer lacks the information of which part of the object it samples
in every step.
Instead, we found that best results could be achieved when not rescaling the conditional information and instead adding
two more tokens to the conditional, specifying top-left and bottom-right position of the current viewport.
During training, we trained on quadratic 2d crops of random size $\in [256, \min(W_{image}, H_{image})]$ and random position.

\subsection{A Note on Evaluation}
\label{subsec:app-evaluation}
\noindent All samples are drawn with temperature $T=1.0$, and a \textit{top\_k}-value of $k=100$ for the conditionals $p(s_i \vert s_{<i})$.

\noindent FID and SceneFID were computed using the Torch-Fidelity package~\cite{obukhov2020_torchfidelity}.
In Table~\ref{tab:cococomparison}, we evaluate our models on 2048 unaugmented test images of the COCO Segmentation
Challenge split to ensure comparability.
These settings remain even if training is done on the full COCO superset.
We do not apply any data augmentation on the test split and only resize to a square (even if during training
we worked on cropped/undistorted images).
In Table~\ref{tab:vgcomparison}, we evaluate on 5088 images of the VG test split in the same way.

\noindent In Table~\ref{tab:allcomparison}, we use slightly different evaluation settings.
In order to ensure comparability of all values in the table, we evaluate on 2048 images of the test split for all three datasets.
Since the models were trained on undistorted random crops of the original images, we do evaluation on center crops of the images.

\subsection{Limitations}
\label{subsec:app-limitations}
\noindent Although our model achieves convincing results both qualitatively and quantitatively, it should be noted that the use of
transformer models also has its price: On the one hand, autoregressive sampling is slow compared to parallel methods,
on the other hand, transformers are known to be data-demanding.
The former can be interpreted as a trade-off between quality and speed.
Regarding the latter, we note that also data augmentation seems to virtually increase the amount of data.
For instance, on COCO, we find that 1d random crops along the longer image axis give performance improvements roughly
equivalent to a $250\%$ increase of training data.
\newpage

\figurezeroshotimages
\figurecomparisonallmethods
\figurereconstruction
\figuresamplehalf
\figurecomparisonlostgan

\end{document}